\documentclass[sigconf]{acmart}

\usepackage{amsmath,amssymb,amsfonts}
\usepackage{algorithmic}
\usepackage{graphicx}
\usepackage{tabularx}
\usepackage{booktabs}
\usepackage{textcomp}
\usepackage{xcolor}
\usepackage{times}
\usepackage{helvet}
\usepackage{courier}
\usepackage {subcaption}
\usepackage{makecell}
\usepackage{multirow}
\usepackage{url}
\usepackage{soul}

\def\BibTeX{{\rm B\kern-.05em{\sc i\kern-.025em b}\kern-.08em
    T\kern-.1667em\lower.7ex\hbox{E}\kern-.125emX}}

\begin{document}

\title{AddrLLM: Address Rewriting via Large Language Model on Nationwide Logistics Data\\}

\author{Qinchen Yang}
\affiliation{%
  \institution{Rutgers University}
  \city{Piscataway}
  \state{New Jersey}
  \country{USA}}
\email{qy129@rutgers.edu}

\author{Zhiqing Hong}
\affiliation{%
  \institution{Rutgers University}
  \city{Piscataway}
  \state{New Jersey}
  \country{USA}}
\email{zhiqing.hong@rutgers.edu}

\author{Dongjiang Cao}
\affiliation{%
  \institution{JD Logistics}
  \city{Beijing}
  \country{China}}
\email{caodongjiang1@jd.com}

\author{Haotian Wang}
\affiliation{%
  \institution{JD Logistics}
  \city{Beijing}
  \country{China}}
\email{wanghaotian18@jd.com}

\author{Zejun Xie}
\affiliation{%
  \institution{Rutgers University}
  \city{Piscataway}
  \state{New Jersey}
  \country{USA}}
\email{zx180@scarletmail.rutgers.edu}

\author{Tian He}
\affiliation{%
  \institution{JD Logistics}
  \city{Beijing}
  \country{China}}
\email{tim.he@jd.com;}

\author{Yunhuai Liu}
\affiliation{%
\institution{Peking University}
\city{Beijing}
\country{China}}
\email{yunhuai.liu@pku.edu.cn}

\author{Yu Yang}
\affiliation{%
  \institution{Lehigh University}
  \state{Bethlehem}
  \state{PA}
  \country{USA}}
\email{yuyang@lehigh.edu}

\author{Desheng Zhang}
\affiliation{%
  \institution{Rutgers University}
  \city{Piscataway}
  \state{New Jersey}
  \country{USA}}
\email{desheng@cs.rutgers.edu}

\begin{abstract}
Textual description of a physical location, commonly known as an address, plays an important role in location-based services(LBS) such as on-demand delivery and navigation.
However, the prevalence of abnormal addresses, those containing inaccuracies that fail to pinpoint a location, have led to significant costs. 
Address rewriting has emerged as a solution to rectify these abnormal addresses. Despite the critical need, existing address rewriting methods are limited, typically tailored to correct specific error types, or frequently require retraining to process new address data effectively.
In this study, we introduce AddrLLM, an innovative framework for address rewriting that is built upon a retrieval augmented large language model. AddrLLM overcomes aforementioned limitations through a meticulously designed Supervised Fine-Tuning module, an Address-centric Retrieval Augmented Generation module and a Bias-free Objective Alignment module.
To the best of our knowledge, this study pioneers the application of LLM-based address rewriting approach to solve the issue of abnormal addresses.
Through comprehensive offline testing with real-world data on a national scale and subsequent online deployment, AddrLLM has demonstrated superior performance in integration with existing logistics system. It has significantly decreased the rate of parcel re-routing by approximately 43\%, underscoring its exceptional efficacy in real-world applications.
\end{abstract}



\maketitle

\section{Introduction}

Addresses are crucial for logistics, ensuring the smooth operation of business by facilitating accurate and efficient delivery processes. In countries like India~\cite{abnormal} and China~\cite{hong2022fastaddr}, the prevalence of inaccurate or abnormal addresses poses a significant challenge. 
This issue arises from inadequate address regulatory frameworks, intricate address structures~\cite{abnormal1}, and click farming fraud~\cite{fraudClick}. 
Abnormal Chinese addresses, defined as those that cannot be parsed into the standard hierarchy (Appendix~\ref{section-address}), often include errors such as missing administrative regions, nested addresses, unofficial aliases, irrelevant words, and misspellings (Appendix~\ref{section-error}). 
An illustrative example is an address that conflates Beijing and Nanjing, which we term a \textit{nested address}. 
These are often exploited for region-specific discounts but lead to unreliable outcomes from Location-Based Services (LBS) due to their lack of systematic recording in databases.

\begin{figure}[h]
    \centering
    \includegraphics[width=\linewidth]{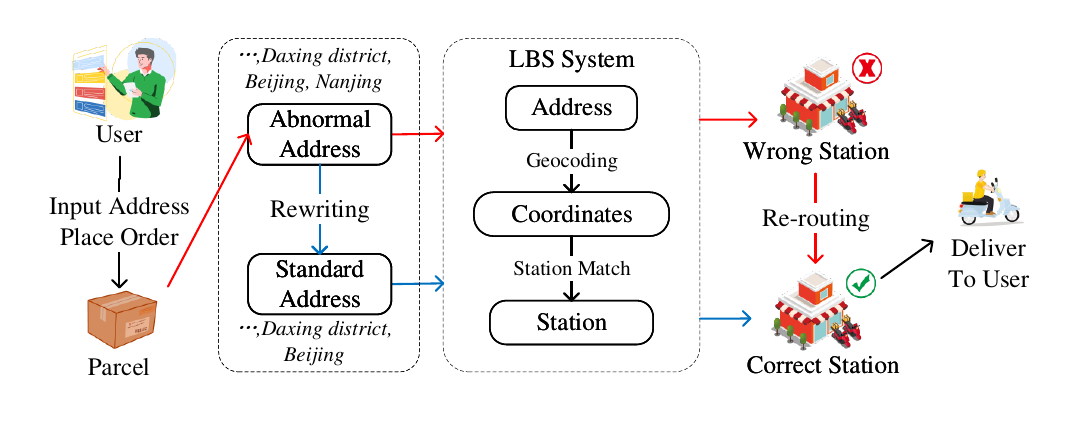}
    \caption{Parcel dispatching, re-routing and address rewriting. Red arrow: abnormal address results in parcel re-routing. Blue arrow: address rewriting prevents re-routing.}
    \label{figure-misclassification}
\end{figure}

This issue significantly impacts companies like JD Logistics, one of the largest logistics companies in the world, which faces around 25,000 daily re-routing events caused by abnormal addresses. 
These misrouted parcels, resulting from addresses that dispatch parcels to the wrong delivery stations, lead to additional transfers and re-routing, depicted in Figure~\ref{figure-misclassification} (red arrow). 
This process incurs annual losses exceeding \$2 million for JD Logistics. 
Address rewriting, a critical procedure that aligns user-submitted addresses with standardized formats, can significantly reduce dispatching errors. 
Empirical evidence suggests that refining user-provided addresses through rewriting, as illustrated in Figure~\ref{figure-misclassification} (blue arrow), can substantially diminish these errors. 
Therefore, developing a comprehensive and adaptable address rewriting framework is imperative to process a wide array of erroneous user address inputs, enhancing the overall efficiency of LBS systems within the logistics sector.

Address rewriting is a special case within the broader field of query rewriting, which has gained prominence and widespread application across e-commerce~\cite{long-tail-taobao,QRICDE,QR4}, question-answering system~\cite{QR3,QR5,QR6}, code search~\cite{li2024rewriting,QR2}. 
Unlike e-commerce queries and natural language questions, addresses posses a more structured format and are closely tied to geographical information. 
Conversely, they are less structured than code and embody more natural language semantics, positioning address rewriting as a distinct category within the spectrum of rewriting tasks.
Due to the close association of address rewriting with business-oriented applications 
such as geocoding, navigation and point-of-interest(POI) matching, which may intersect with proprietary algorithms at the heart of related companies' main operations, there is a scarcity of published studies in this particular area. 
Like query rewriting, existing literature on address rewriting can be classified into approaches based on Statistical Machine Translation(SMT) or Neural Machine Translation(NMT). SMT-based methods leverage statistical models to transform an original address into its modified counterpart. Such methods often encounter performance constraints due to the limited representation capabilities of statistical models as observed in the early Bing Maps address rewriting system outlined in ~\cite{roy2023deep}.
NMT-based methods~\cite{liu2021geo,roy2023deep} employ an encoder-decoder architecture where the encoder maps original address into a latent representation, and the decoder translates this representation into the modified address. Furthermore, some works~\cite{kakkar2018address,tong2022context} focus on extracting related information for address rewriting, which can be viewed as prerequisite task of address rewriting.
However, ~\cite{liu2021geo,roy2023deep} require re-training when new addresses are added into database, a common occurrence in the industry. ~\cite{liu2021geo} necessitates explicit geographical knowledge for generating meaningful embeddings, which may not always be readily available. ~\cite{roy2023deep,kakkar2018address,tong2022context} are constrained by its focus on particular types of address anomalies, such as alternations, spell mistakes and alias, which represent only a fraction of the errors encountered real-world data stream, as illustrated in Appendix~\ref{section-error}. Consequently, there is a pressing need for a robust framework for address rewriting that can operate effectively with the textual address data alone, handle adequate errors of address in a unified model and does not necessitate constant retraining with the addition of new addresses.

Retrieval-augmented Large Language Model is a potential solution to the constrained scope problem and retraining problem. Firstly, with the development of LLM techniques\cite{GPT-4,hadi2023survey,chang2024survey}, reasoning capability of LLM shows the potential to rectify extensive errors of addresses, even those never considered in previous works(Table~\ref{table-address}, in a unified model.
Secondly, LLM itself has the same retraining problem as previous works. To solve this, Retrieval Augmented Generation(RAG) decouples knowledge storage and reasoning capability of LLM~\cite{huang2024survey,gao2023retrieval}, making it possible to generate updated answers without retraining LLM, when new knowledge appears. 

However, utilizing RAG-based LLM to solve address rewriting problem is still challenging, because: (1) current public LLMs lack address comprehension and related knowledge; (2) current RAG frameworks are mainly designed for question-answering tasks and natural language paragraphs data, whose semantics diverge far from addresses; (3) current objective alignment methods are mainly based on trainable reward model, which involves additional training efforts and may introduce bias and inaccuracies.

To solve these challenges, in this paper, we propose AddrLLM, an address rewriting framework based on retrieval augmented large language model. We build a RAG framework specifically for address data, perform supervised fine-tuning for LLM on logistics related tasks, and design a bias-free and large-scale objective alignment framework based on JD's LBS system.

In summary, the main contributions of this work are as follows:
\\\noindent$\bullet$~We are the first to explore the possibility of utilizing retrieval augmented large language model to rewrite address and subsequently deploy in real-world data stream.
\\\noindent$\bullet$~We design a novel LLM-based framework, AddrLLM, integrating multi-instruction supervised fine-tuning, bias-free objective alignment and address-centric retrieval augmented generation modules, to solve challenges associated with adopting LLM to address rewriting.
\\\noindent$\bullet$~By offline experiments, AddrLLM rectifies 43.9\% abnormal addresses, and outperforms SoTA methods by 24.2\%. AddrLLM has been deployed at JD Logistics in Zhejiang province for over four months, and effectively reduced over 40\% parcel re-routing caused by abnormal addresses among around 2 million daily parcels.

\section{Method}~\label{section-method}
\begin{figure}[ht]
    \centering
    \includegraphics[width=\columnwidth]{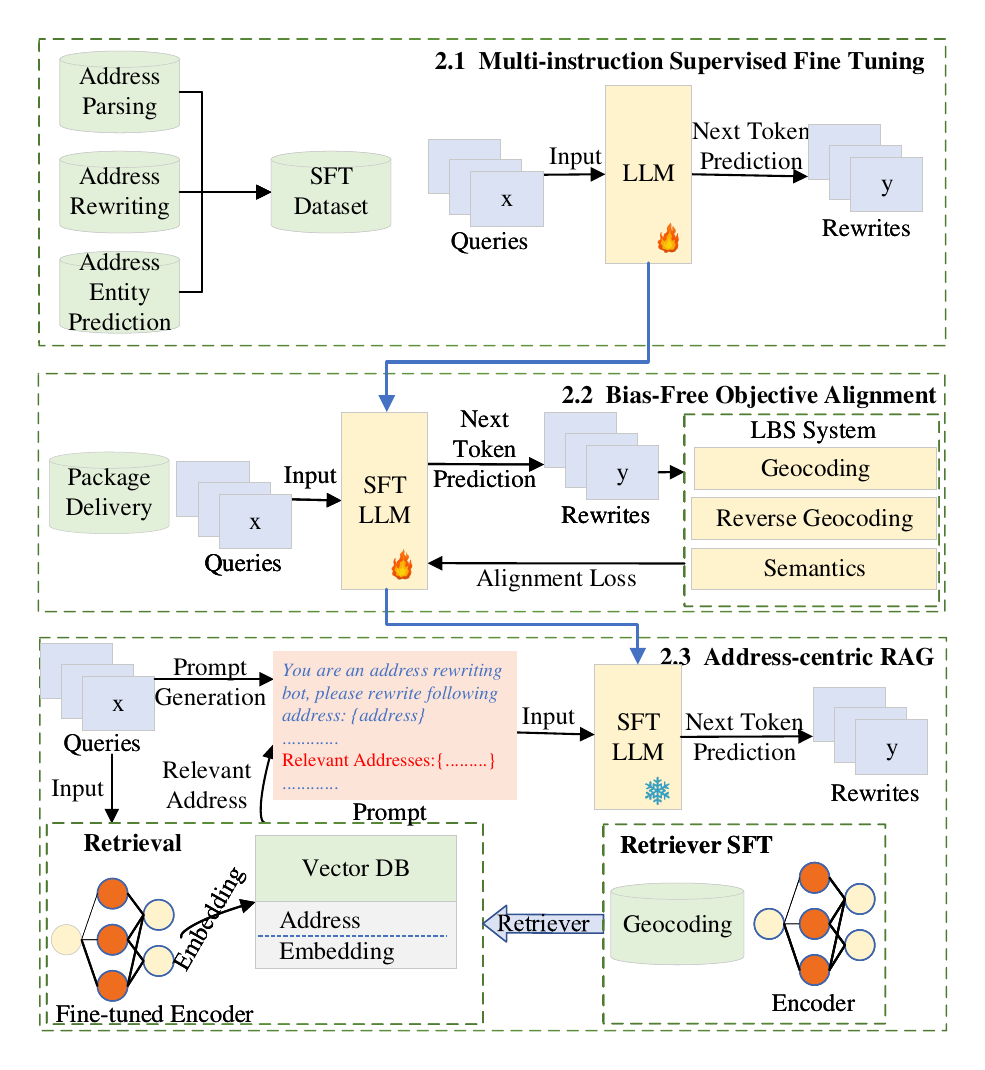}
    \caption{Framework of AddrLLM}
    \label{figure-model}
\end{figure}

Address rewriting aims to refine user-input addresses into a standardized format that aligns with the user's original intent. We introduce a comprehensive framework for LLM-based address rewriting, AddrLLM, depicted in Figure~\ref{figure-model}. AddrLLM is composed of three key components: Supervised Fine-tuning(SFT) module, Address-centric Retrieval Augmented Generation(RAG) module and Bias-free Objective Alignment module.
Initially, we leverage JD's sophisticated Location-Based Services(LBS) system to collect an extensive, high-quality dataset specifically for the SFT task associated with address rewriting.
Subsequently, we design the objective alignment module to further calibrate the generation of rewrites to desired results. To prevent potential bias arising from reward model or manual annotation, we integrate the LBS system, which provides bias-free feedback that is directly derived from the model's performance in rewriting task.
Finally, to enhance the rewriting process, we develop a customized RAG module designed to enrich the LLM with contextual information via targeted retrieval of relevant addresses.

\begin{table}[ht]
    \centering
    \caption{Details of datasets}
    \begin{tabularx}{\linewidth}{c|c|c}
    \hline
    Usage&Size&Task\\
    \hline
    Retriever Fine-tuning&200M&Geocoding\\
    \hline
    \multirow{3}*{LLM Fine-tuning}&20M&Address Parsing\\
    &20M&Address Entity Prediction\\
    &20M&Address Rewriting\\
    \hline
    LLM Objective Alignment&4M&Package Delivery\\
    \hline
    \multirow{3}*{Model Testing}&0.5M&Address Entity Prediction\\
    &0.5M&Address Rewriting\\
    &0.5M&Geocoding\\
    \hline
    \end{tabularx}
    \label{table-datasets}
\end{table}

\subsection{Multi-instruction Supervised Fine-tuning}
Because the semantics of address significantly diverges from the pre-training corpus of LLMs, directly employing these models for address processing may result in inaccuracies. To solve this challenge, we adopted a strategy that involves aggregating a range of tasks related to address to fine-tune LLMs, thereby improving their proficiency in understanding standard Chinese addresses. Here we describe the tasks and datasets for SFT. The prompt design for these tasks are detailed in Appendix~\ref{section-prompt}.

\noindent$\bullet$~\textbf{\textit{Address Parsing}}: Address parsing task involves the process of breaking down an address into its constituent components. The structure of a Chinese address is detailed in Appendix~\ref{section-address}. Our address parsing dataset comprises 20 million <address, address components> pairs obtained from JD's LBS system. Fine-tuning LLMs on the address parsing task is likely to help these models understand the structure of standard addresses, enabling them to generate addresses that conform to the standard hierarchy.

\noindent$\bullet$~\textbf{\textit{Address Entity Prediction(AEP)}}: Address Entity Prediction is to infer absent administrative region in address.
In logistics system, user-provided addresses frequently lack administrative regions, posing significant challenges to package dispatching. 
Thus, address entity prediction or filling is an important ability of address rewriting model.
For the AEP task, we collect 20 million addresses from historical delivery orders in JD, and randomly delete administrative regions. Then the address rewriting model needs to predict the missing address entity.

\noindent$\bullet$~\textbf{\textit{Address Rewriting}}: Our address rewriting dataset is generated from two sources. 
The primary source is the JD LBS system.
When a user inputs an address, the LBS system may do some basic rewriting, such as misspelling correction. We collect user-input addresses and rewritten ones and obtain 15 million samples from LBS system.
The second source is address recommendation system integrated on JD e-commerce platform.
When a user places an order with delivery address on JD e-commerce platform, the system recommends some related addresses from our standard address database. If user chooses to replace original address with a recommended address, we keep a record. However, this data can be noisy because the original address provided by user may only contain some keywords. Thus, we filter records by feeding original addresses into JD LBS system, and if geocoding succeeds(geocoding doesn't return error although returned coordinates may be incorrect), we can view original address as a complete address and put corresponding record into our address rewriting dataset. Following this process, 5 million samples are collected by the second source. In total, our address rewriting dataset contains 20 million samples. This dataset includes 77.8\% samples that do not involve any rewriting, i.e. original address and rewritten address are the same.

Finally, address parsing, address entity prediction and address rewriting datasets are mixed together to form the SFT dataset $D$.

\noindent$\bullet$~\textbf{\textit{Multi-instruction Supervised Fine-tuning (SFT)}}: The process of generating text using large language model can be viewed as auto-regressive sampling. In auto-regressive language generation, each word is predicted one at a time, and each prediction conditions on the prompt and previously generated words. Given model input (prompt) x and standard output y, the training objective is to find the model parameters $\theta^*$ that maximizes the conditional probability $p(y|x)=\prod_{i=1}{p(y_i|y_{0:i-1},x)}$. The training objective can be formulated as:
\begin{equation}
\setlength{\abovedisplayskip}{2pt}
\setlength{\belowdisplayskip}{2pt}
    \theta^* = \arg\max_{\theta} \sum_{(x, y) \in D} - \log \pi(y_i|y_{0:i-1},x;\theta)
\end{equation}
where $D$ is the SFT dataset, $\pi$ is our large language model and $\theta$ is its parameters.

\subsection{Bias-free Objective Alignment}\label{section-objectiveAlignment}
Through prior multi-instruction SFT, AddrLLM has developed understanding of standard Chinese addresses.
Because address rewriting SFT dataset is generated based on JD LBS system, i.e. these abnormal addresses can be identified and rectified by current system, it indicates that these abnormal addresses have different distribution from those cannot be rectified by LBS system. 
As a result, AddrLLM's current skill set does not extend to effectively rewriting addresses that fall outside the correction capabilities of the JD LBS system. Moreover, the address groundtruth of unsuccessful delivery events is unknown to us. Therefore, to further enhance address rewriting capability of AddrLLM, particularly for those challenging addresses beyond the reach of the JD LBS system, reinforcement learning based objective alignment is necessary.

In previous works, frameworks for aligning the objectives of LLM typically rely on either manual annotations or trainable reward model~\cite{rewardModel1,rewardmodel2,rewardmodel3} to guide LLM's training. However, these methods have some limitations that hinder their application in industrial settings. Firstly, manual annotations are not only biased, but also expensive to extend to large scale data. Additionally, the performance of LLM is highly tied to the effectiveness of reward model training, which produces scores that are not easily interpretable.

To overcome these limitations, we introduce a novel approach that utilizes JD LBS system for providing feedback. Our approach provides an unbiased and fast reward calculation that is directly tied to LLM's performance on address rewriting task. More importantly, the reward calculation is interpretable, which is vital for robust applications and further improvements in the industry. The reward $r$ is comprised of three scores: semantics score, reverse geocoding score and geocoding score. 

Firstly, the rewritten address should not be too far from the original address in semantics. For example, if user inputs "Outlets Store (Eastern door)", the model should not simply rewrite it to "Outlets Store", even if the geocoding results of two addresses are close spatially. To prevent rewriting from deviating too far from user's intent, we design semantics score $seman(x,y)$:
\begin{equation}
\setlength{\abovedisplayskip}{2pt}
\setlength{\belowdisplayskip}{2pt}
    seman(x,y) = cos(f(x),f(y))
\end{equation}
where $x$ is original address, $y$ is rewritten address, $cos$ is the function of cosine similarity, formulated as $cos(\vec{A},\vec{B})=\frac{\vec{A} \cdot \vec{B}}{|\vec{A}| |\vec{B}|}$, $f$ is semantics embedding model. At experiments, we choose pre-trained model BERT as $f$.

Secondly, the rewritten address should be semantically close to the address obtained by reverse-geocoding on delivery coordinates. The result of reverse-geocoding may not be exact, for example, the room number or building number may be incorrect. However, the rewritten address should be semantically close to that address to some extent, for example, they are in the same community or road. To directly measure the rewritten address, we design reverse geocoding score $revgeo(y,c)$, formulated as:
\begin{equation}
\setlength{\abovedisplayskip}{2pt}
\setlength{\belowdisplayskip}{2pt}
    revgeo(y,c) = cos(f(y),f(reverse(c)))
\end{equation}
where $y$ is rewritten address, $c$ is coordinates of successful delivery, $reverse$ is reverse-geocoding service, $f$ is semantics embedding model.

Thirdly, we choose geocoding task to evaluate the rewriting result. Intuitively, the geocoding result of rewritten address should be close to the coordinates of successful delivery. Thus, we design geocoding score $geo(y,c)$, formulated as:
\begin{equation}
\setlength{\abovedisplayskip}{2pt}
\setlength{\belowdisplayskip}{2pt}
    \begin{aligned}
    &geo(y,c) = 
    \begin{cases}
        1          & k<\theta_1 \\
        1-(k-\theta_1)/\theta_2 & \theta_1 \leq k < \theta_2 \\
        0          & k \geq \theta_2 \\
        0          & geocoding\ failure\\
    \end{cases},\\
    &k = dis(geocoding(y),c)
    \end{aligned}
\end{equation}
where $dis$ is Euclidean distance function, $geocoding$ is JD geocoding system that maps from address to coordinates. We normalize geocoding score to [0,1] by weights $\theta_1$ and $\theta_2$. In experiments, we set $\theta_1$ as 100 meters and $\theta_2$ as 1000 meters. When the rewritten address cannot be recognized by JD geocoding service as an address (geocoding failure), the geocoding score is 0.

Finally, three scores are added together with weights $\lambda_1$,$\lambda_2$,$\lambda_3$:
\begin{equation}
\setlength{\abovedisplayskip}{1pt}
\setlength{\belowdisplayskip}{1pt}
    r(x,y,c) = \lambda_1 seman(x,y) + \lambda_2 revgeo(y,c) + \lambda_3 geo(y,c)
\end{equation}
In experiments, $\lambda_1$,$\lambda_2$ and $\lambda_3$ are set as 0.2,0.2 and 0.6 respectively.

\noindent~\textbf{RL Task Formulation}: 
At each time step $t$, large language model $\pi$ generate next token as action $a_t$, based on current state $s_t$, which includes already generated tokens. Then the model obtains an immediate reward $r_t$ by a rewarding function $R:S\times A\rightarrow \mathbb{R}$.
The detailed Markov Decision Process(MDP) formulation is in Appendix~\ref{section-rl}.

\noindent\textbf{\textit{Training}}:
We adopt Proximal Policy Optimization(PPO)~\cite{schulman2017proximal} to optimize the large language model. 
The PPO algorithm can be formulated as:
\begin{equation}
\setlength{\abovedisplayskip}{1pt}
\setlength{\belowdisplayskip}{2pt}
\begin{aligned}
    &\max_{\theta} \mathbb{E}_{(s_t,a_t)\sim\pi_{\theta'}}[\min \{ k_{t,\theta}A^{\theta'}(s_t,a_t),\\
    &\hspace{1cm} clip(k_{t,\theta},1-\epsilon,1+\epsilon)A^{\theta'}(s_t,a_t) \} ]\\
    &k_{t,\theta} = \frac{p_\theta(a_t|s_t)}{p_{\theta'}(a_t|s_t)}
\end{aligned}
\end{equation}
where $\theta'$ is parameters of fixed policy, $\theta$ is parameters of updated policy, the clip function $clip(k_{t,\theta},1-\epsilon,1+\epsilon)$ limits the ratio $k_{t,\theta}$ to the range$[1-\epsilon,1+\epsilon]$. $A$ is advantage function, which is formulated based on the estimation of value network $V_\phi$. The value network $V_\phi$ is initialized from the policy network $\pi_0$. The formulation follows Generalized Advantage Estimation(GAE)~\cite{schulman2015high}:
\begin{equation}
\setlength{\abovedisplayskip}{2pt}
\setlength{\belowdisplayskip}{1pt}
\begin{aligned}
    &\delta_t = R(s_t,a_t) + V_\phi(s_{t+1}) - V_\phi(s_t)\\
    &A_t^{\theta}(s_t,a_t) = \sum_{t'=0}^{\infty}\lambda^{t'}\delta_{t+t'}
\end{aligned}
\end{equation}
where $\lambda$ is the bias-variance trade-off parameter.

To prevent the model from deviating too far form the initialization, we add a KL-divergence regularization to reward~\cite{ramamurthy2022reinforcement}:
\begin{equation}
\setlength{\abovedisplayskip}{2pt}
\setlength{\belowdisplayskip}{2pt}
    R(s_t,a_t) = r(x,y,c) - \beta KL(\pi_\theta||\pi_0)
\end{equation}

The final loss function is composed of policy loss and value loss:
\begin{equation}
\setlength{\abovedisplayskip}{2pt}
\setlength{\belowdisplayskip}{2pt}
\begin{aligned}
    &\mathcal{L}_\theta = -\frac{1}{|S|T}\sum_{\tau\in S}\sum_{t=0}^{T} \min \{ k_{t,\theta}A^{\theta'}(s_t,a_t),\\
    &\hspace{1cm} clip(k_{t,\theta},1-\epsilon,1+\epsilon)A^{\theta'}(s_t,a_t) \}\\
    &\mathcal{L}_\phi = \frac{1}{|S|T}\sum_{\tau\in S}\sum_{t=0}^{T}(V_\phi(s_t)-R_t)^{2}\\
    &\mathcal{L}_{PPO} = \mathcal{L}_\theta + \lambda_v \mathcal{L}_\phi
\end{aligned}
\end{equation}
where $S$ is sampling set, $T$ is step numbers.

For the objective alignment, we utilize 4 million <address, location> samples, where address is the user-input address and location is the delivery coordinates reported by courier. The groundtruth address is unknown to us. The objective alignment module guides LLM to learn how to rewrite these user-input addresses to standard ones.

\subsection{Address-centric RAG}\label{section-RAG}
Large Language Models (LLMs) often experience hallucination, particularly when generating content in domains unfamiliar to them, as highlighted in the studies by~\cite{YaoZYDSN023,BangCLDSWLJYCDXF23,gao2023retrieval}. Given that LLMs are trained on datasets encompassing a broad range of scenarios, their expertise in specific tasks such as Chinese address rewriting is limited. Consequently, this make them prone to generating hallucinated content when tasked with rewriting Chinese addresses. Meanwhile, LLMs also suffer from misalignment~\cite{lewis2020retrieval,luu2021time}, which is also significant in address system since address database keeps updating. To solve these challenges, we develop an Address-centric Retrieval-Augmented Generation(RAG) module, which decouples reasoning ability and address knowledge storage of LLM, and maintains knowledge in external database.

In this section, we introduce the popular "retrieve-then-read" RAG pipeline and adapt it to our address rewriting scenario. Firstly, the retriever identifies and extracts a set of relevant addresses from the database. Subsequently, the generator, i.e. a fine-tuned LLM, bases its generated output on the addresses retrieved. We describe the retriever in more detail below.

\noindent$\bullet$~\textbf{\textit{Retriever}}: The retriever's function is to identify and prioritize all addresses relevant to the input. Formally, the retriever's role is encapsulated by the function $M$, which maps a query $q$ and a database $K$ to a subset $K'$, such that $M: [q, K] \rightarrow K'$, where $K'$ comprises the relevant addresses from $K$ corresponding to the input $q$, ordered by decreasing relevance. In our scenario, $q$ is an address we want to modify, $K'$ is a set of addresses relevant to $q$. The foundation model of retriever is an encoder model $E$, which maps an address to a representation. Then a similarity score is computed between query address $q$ and each sample in $K$:
\begin{equation}
\setlength{\abovedisplayskip}{2pt}
\setlength{\belowdisplayskip}{2pt}
    s(p,q) = sim(E(q),E(p)), p\in K
\end{equation}
where $sim:\mathbb{R}^d\times\mathbb{R}^d\rightarrow\mathbb{R}_+$ is the similarity function, such as cosine similarity. Finally, samples with highest similarity score, i.e. $K'$, are returned by retriever.

For the sake of efficiency and scalability, retriever usually encode textual information into embedding space, where the retriever performs search~\cite{gao2023retrieval,huang2024survey}. In previous RAG frameworks which mainly target at question-answering task, relevance is defined on semantics of paragraphs and sentences. For our address scenario, however, semantics relevance could introduce misleading information. Our objective necessitates a shift in the relevance criterion towards geographical proximity, ensuring that the retrieved addresses are within a close spatial range to the query address.

\noindent$\bullet$~\textbf{\textit{Spatial Encoding}}: Previous research has explored the spatial encoding of addresses~\cite{huang2022ernie,wu2023g2ptl}. However, these techniques prioritize other NLP tasks like Masked Language Modeling and Hierarchical Text Classification, which generalize the pre-trained model(PTM) at the expense of its capacity for spatial encoding. Thus, fine-tuning the PTM is necessary.
In our retrieval framework, we employ the widely adopted BERT encoder architecture~\cite{Devlin2019BERT} as foundation model $E$. The initialization of our retriever leverages parameters from the text encoder of G2PTL~\cite{wu2023g2ptl}. Subsequently we fine-tune BERT on geocoding task. To achieve this, we append several fully connected(FC) layers to existing BERT structure, which are designed to transform the embedding to geographic coordinates, namely longitude and latitude. Our geocoding dataset comprises 200 million <address, coordinates> pairs. Given that BERT comes pre-trained while the added FC layers are randomly initialized, we adopt a two-phase training strategy. In the first epoch of training, we freeze BERT parameters and concentrate on training FC layers. For subsequent epochs, we train the whole neural network, including BERT and FC layers. When retrieving addresses, the representation produced by BERT, a 768-element vector, is utilized as the spatial embedding.

\section{Experiments}
\noindent In this section, we conduct experiments to answer the following research questions:
\\\noindent$\bullet$~\textbf{\textit{RQ1}}: Does AddrLLM outperform other SoTA methods in offline experiments?
\\\noindent$\bullet$~\textbf{\textit{RQ2}}: Whether and how often does AddrLLM rewrite a standard address to incorrect one?
\\\noindent$\bullet$~\textbf{\textit{RQ3}}: How the components in AddrLLM contribute to the performance?
\\\noindent$\bullet$~\textbf{\textit{RQ4}}: How long is the duration required for SFT and objective alignment of LLM, and the responsiveness of our framework?
\\\noindent$\bullet$~\textbf{\textit{RQ5}}: Does the retriever in RAG module perform well?
\\\noindent$\bullet$~\textbf{\textit{RQ6}}: What improvements has the deployment of AddrLLM brought to JD's LBS system?

\subsection{Testing Datasets and Evaluation Metrics}
In this section, we present the downstream applications that served as benchmarks for assessing our model, along with the associated metrics. The prompts employed for the LLMs across these applications are detailed in Appendix~\ref{section-prompt}. An overview of the datasets for the different tasks is provided in Table~\ref{table-datasets}. While the training datasets have been extensively described in Section~\ref{section-method}, in this part, we focus exclusively on the testing datasets. To test our model, we build three extensive datasets, each corresponding to one of the three applications.\\
\noindent~\textbf{\textit{Address Entity Prediction}}: Request the model to generate addresses with all necessary elements for those lacking administrative regions.\\
$\bullet$~\textbf{Trigger Prediction}: This metric evaluates the ability of the model to discern which input addresses are missing administrative regions and require completion. If the model attempts to complete the address (e.g. add some words), we identify it as successful trigger prediction, regardless of whether the completion is correct.\\
$\bullet$~\textbf{Accuracy}: Whether the model generates correct address. This metric is calculated as the ratio of accurately rewritten addresses to the total number of addresses missing regions.\\
\noindent~\textbf{\textit{Direct}}: Directly evaluate the rewritten address.\\
$\bullet$~\textbf{Hit Rate}: We derived the groundtruth for levels 1 to 4 (Appendix~\ref{section-address}) by reverse geocoding the delivery coordinates, complementing them with level 5 and 6 from the input addresses. After tokenizing both the rewritten addresses and groundtruth addresses, we then assessed the accuracy by calculating the percentage of address components predicted by the model that correspond with the groundtruth components.\\
\noindent~\textbf{\textit{Geocoding}}: Geocoding is a popular GIS service that maps textual address to geospatial coordinates. In JD's LBS system, parcel dispatching relies heavily on the geocoding service. Here we rewrite the input address by model before feeding it to JD's geocoding service of LBS system.\\
$\bullet$~\textbf{Acc@300m}: The percentage of coordinates that locates within 300-meter radius of the groundtruth coordinates, with the groundtruth being the delivery coordinates as reported by couriers.\\
$\bullet$~\textbf{Acc@500m}: A metric akin to Acc@300m, except that the 300-meter radius is substituted with a 500-meter radius.\\
$\bullet$~\textbf{Acc@Station}: This metric calculates the percentage of coordinates that locates within the responsible area of groundtruth delivery station.
In logistics, if an address is correctly located within the area of corresponding station, we can view it as a successful geocoding process, because subsequently couriers in station do not relay on precise coordinates but textual addresses to perform last-mile delivery~\cite{lastMileDeliverySurvey}.\\
$\bullet$~\textbf{Robustness}: Our geocoding testing dataset contains 90\% addresses that can be correctly identified by JD's LBS system. In industrial scenario, an important metric is that the model should not rewrite a correct address to incorrect address. Thus, we design the metric Robustness, which is calculated as the percentage of originally correct addresses remaining correct after rewriting.\\
$\bullet$~\textbf{Correction}: This metric measures the percentage of abnormal addresses that can be rectified by mode.\\
The datasets used for direct evaluation and geocoding contain an identical collection of addresses. Within this set, JD's LBS system accurately dispatches 90\% of the addresses to delivery station, while the remaining 10\% are subject to re-routing. Addresses within testing datasets do not overlap with those in training datasets.

\subsection{Implementation Details}
We choose AdamW~\cite{AdamW} optimizer for LLM SFT and objective alignment. During SFT stage, LLM is trained for 1 epoch with learning rate 1e-5. During objective alignment stage, LLM is trained for 4 epochs with learning rate 1e-6. During retriever fine-tuning stage, BERT is trained for 4 epochs with learning rate initialized to 5e-5 and gradually decreased during the process of training, where Adam~\cite{Adam} is the optimizer. We compare Qwen-7B~\cite{qwen} and Baichuan-7B~\cite{baichuan7b} as base LLM because of their outstanding performance in Chinese language related tasks. At retrieval stage, retriever select top-10 most relevant addresses from database, for which we use the open source vector database Vearch~\cite{li2019design}. All of the offline experiments are conducted on a cloud-based computational cluster including 10 Nvidia H800 GPUs, 160 CPU cores of Intel Xeon 8468V.

\newcolumntype{Y}{>{\centering\arraybackslash}X}
\begin{table*}[ht]
    \centering
    \caption{Offline Experiment Result}
    \begin{tabularx}{0.95\textwidth}{l|cc|c|ccccc|Y}
    \hline
         \multicolumn{1}{c|}{\multirow{2}*[-2mm]{Method}}&
         \multicolumn{2}{c|}{\textbf{AEP}}&
         \textbf{Direct}&
         \multicolumn{5}{c|}{\textbf{Geocoding}}&\multirow{2}*[-2mm]{\textbf{Average}}\\
         &\makecell[c]{Trigger \\Prediction}$\uparrow$&Acc$\uparrow$&Hit Rate$\uparrow$&\makecell[c]{Acc@\\300m}$\uparrow$&\makecell[c]{Acc@\\500m}$\uparrow$&\makecell[c]{Acc@\\Station}$\uparrow$&Robustness$\uparrow$&Correction$\uparrow$\\
         \hline
         BART              &51.2   &46.9   &55.9   &59.4&  63.8    &62.1   &68.8   &1.8&51.2\\
         BERT              &57.3   &47.2   &56.7   &63.5   &69.5   &64.3   &71.2   &2.3&54\\
         G2PTL             &\underline{84.8}   &\underline{79.5}   &\underline{74.3}   &\underline{77.4}   &\underline{79.2}   &\underline{87.2}   &94.7   &\underline{19.7}&\underline{74.6}\\
         QWen-7B           &70.6   &64.3   &68.7   &73.5   &76.2   &79.1   &86.9   &8.4&65.9\\
         Baichuan-7B       &71.4   &65.8   &69.2   &75.7   &78.9   &79.8   &87.5   &10.2&67.3\\
         SoP               &-      &-      &-      &88.3   &89.1   &90.0   &\textbf{100}  &0&-\\
         \hline
         \textbf{AddrLLM(ours)}     &\textbf{91.8}   &\textbf{90.3}   &\textbf{89.7}   &\textbf{91.6}   &\textbf{92.7}   &\textbf{94.3}   &\underline{99.9}   &\textbf{43.9}&\textbf{86.8}\\
         -AddrLLM-QWen      &89.9   &88.3   &86.7   &89.4   &90.6   &93.8   &99.6   &41.4&84.9\\
         -AddrLLM w/o OA            &88.5   &84.6   &83.8   &85.7   &87.1   &88.8   &96.2   &21.8&79.6\\
         -AddrLLM w/o SFT           &76.3   &71.7   &74.8   &78.2   &82.3   &84.5   &91.2   &24.6&72.9\\
         -AddrLLM w/o RAG           &85.6   &82.4   &79.6   &81.6   &83.1   &87.5   &93.5   &33.5&78.4\\
         -AddrLLM-RAG               &74.1   &69.8   &71.7   &77.3   &80.6   &81.8   &89.2   &14.9&69.9\\
    \hline
    \end{tabularx}
    \label{table-results}
\end{table*}

\subsection{Compared Methods}
\noindent$\bullet$~\textbf{\textit{BART}}~\cite{BART}: A widely used transformer-based encoder-decoder Pre-Trained Model(PTM), which achieve remarkable gains in NLP tasks. We fine-tune it on our address rewriting dataset to enhance its ability in rewriting addresses.
\\\noindent$\bullet$~\textbf{\textit{BERT}}~\cite{Devlin2019BERT}: A popular transformer-based encoder PTM. We append Transformer decoder and fine-tune it on address rewriting dataset.
\\\noindent$\bullet$~\textbf{\textit{G2PTL}}~\cite{wu2023g2ptl}: The latest PTM pre-trained on logistics data and tasks. The built-in Address Entity Prediction module in G2PTL needs missing words marked as "$[MARK]$", which cannot be directly applied on our address rewriting scenario. Thus, we utilize the text encoder of G2PTL and append transformer decoder to return rewritten address. We fine-tune it on address rewriting dataset.
\\\noindent$\bullet$~\textbf{\textit{QWen-7B}}~\cite{qwen}: LLM baseline. Prompt Qwen-7B to rewrite address.
\\\noindent$\bullet$~\textbf{\textit{Baichuan-7B}}~\cite{baichuan7b}: LLM baseline. Prompt Baichuan-7B to rewrite address.
\\\noindent$\bullet$~\textbf{\textit{SoP}}: Address geocoding service within JD's LBS system.
\\\noindent$\bullet$~\textbf{\textit{AddrLLM-Qwen}}: Utilize QWen-7B as the base LLM instead of Baichuan-7B.
\\\noindent$\bullet$~\textbf{\textit{AddrLLM w/o OA}}: Remove the objective alignment module from AddrLLM.
\\\noindent$\bullet$~\textbf{\textit{AddrLLM w/o SFT}}: Remove the SFT module from AddrLLM.
\\\noindent$\bullet$~\textbf{\textit{AddrLLM w/o RAG}}: Remove the RAG module from AddrLLM.
\\\noindent$\bullet$~\textbf{\textit{AddrLLM-RAG}}: Remove SFT and objective alignment modules from AddrLLM.

\subsection{Offline Experiment Result}
Our experimental results on direct evaluation, address entity prediction and geocoding tasks are shown in Table~\ref{table-results}.
By analyzing the results, we have the following findings:

\subsubsection{Overall Performance (\textbf{\textit{RQ1}})}
Upon evaluating the accuracy of AEP, the hit rate and geocoding accuracy, it becomes evident that our model, AddrLLM, outperforms all of baselines on our testing dataset, which consists of 90\% standard and 10\% abnormal addresses. The SoTA baseline G2PTL achieves second best average performance.
Pose the application of AddrLLM for address rewriting, compared with the second best model G2PTL, we observed significant improvements across several metrics: trigger prediction of AEP increases by 7\%, the accuracy of AEP increases by 10.8\%, the hit rate increases by 15.4\%, the geocoding accuracy at station level increases by 7.1\%. 
This enhancements underscore the efficacy of our LLM-based address rewriting framework in detecting and correcting address inaccuracies. Notably, in the context of geocoding, a critical factor for successful parcel delivery in the logistics industry, compared with SoP method, AddrLLM reduces station-level inaccuracies by a substantial 43\%.

\subsubsection{Robustness (\textbf{\textit{RQ2}})}
Since in the real-world scenario, percentage of incorrect addresses is greatly lower than that of our testing dataset(10\%), it is important to test whether model rewrites a standard address to incorrect one. Thus, we design robustness metric on geocoding task. Robustness measures the percentage of standard addresses remaining correct after rewriting. From the table, AddrLLM achieves 99.9\% robustness, which significantly outperforms other baselines. This suggests that AddrLLM is effective at recognizing correct addresses and avoids modifying them into incorrect ones.

\subsubsection{Ablation Study (\textbf{\textit{RQ3}})} To understand the contribution of each component to AddrLLM's performance and to inform future enhancements and deployment strategies, we conduct an ablation study and meticulously analyze the experimental results. Our analysis yields the following insights:

The original Baichuan and QWen models exhibit suboptimal performance in address rewriting tasks, falling short of the SoTA baseline established by G2PTL, by an average of 7.3\% and 8.7\% respectively. This indicates that generalized LLMs have limited knowledge about standard addresses. In contrast, compared with Baichuan and QWen, AddrLLM and AddrLLM-QWen obtain improvements by 19.5\% and 19\%, respectively. This indicates that the integration of our novel RAG, SFT and objective alignment modules significantly enhance LLM's capacity for address rewriting.

In the absence of objective alignment(OA) component, AddrLLM attains the highest performance relative to other ablated models except AddrLLM-QWen. Even when OA is removed, AddrLLM still outshines the SoTA baselines, albeit it falls behind the SoP in geocoding task. The SFT process relies on groundtruth data generated by JD's LBS system(SoP). Therefore, without the OA module, it is challenging for AddrLLM to surpass the SoP. The OA module, however, introduces samples where SoP lacks the knowledge for the groundtruth addresses, providing a unique training opportunity for AddrLLM to refine its performance and ultimately exceed the SoP.

Excluding the Supervised Fine-Tuning (SFT) module leads to a marked reduction in AddrLLM's effectiveness, particularly affecting the robustness of geocoding task. This deterioration is due to the prevalence of non-rewriting samples within the address parsing, rewriting and entity prediction datasets of SFT stage. These non-rewriting samples are instrumental in enhancing the model's proficiency in recognizing correct addresses and determining when not to perform rewrites. 

When the RAG component is removed, there is a noticeable downturn in AddrLLM's effectiveness, indicating the integral role of RAG in the model's overall functionality.

Observing AddrLLM-RAG, where we remove SFT and OA modules, our model achieves the lowest performance relative to other ablated models. This is because the original Baichuan model lacks understanding of standard addresses. The average performance is only 2.6\% higher than that of Baichuan-7B. This indicates that without fine-tuning the base model, AddrLLM is hard to utilize the relevant addresses retrieved by RAG module.

In conclusion, the synergy of SFT, OA, and RAG is essential for optimizing AddrLLM in the specialized task of rewriting addresses.

\subsubsection{Complexity Analysis (\textbf{\textit{RQ4}})}
In Table~\ref{table-time}, we present a consolidated overview of the cumulative duration, measured in hours, required for Supervised Fine-Tuning (SFT), objective alignment, and evaluation phases. Notably, the objective alignment stage operates at an average rate of 13 samples per second. This relatively slow speed is attributed primarily to the reward calculation being executed on the CPU, whereas the training of the LLM occurs on the GPU. Consequently, there is an increased frequency of data transfers between the CPU and GPU compared to typical neural network training procedures. Additionally, this transfer and training process is inherently difficult to parallelize, further contributing to the reduced processing speed.

\begin{table}[h]
    \centering
    \caption{Complexity Analysis}
    \begin{tabular}{c|c|c|c}
    \hline
    Stage&Task&\makecell[c]{Average Speed\\(\#sample/s)}&\makecell[c]{Total\\(hours)}\\
    \hline
    \multirow{3}*{SFT}  &Parsing    &51     &109\\
                        &AEP        &79     &71\\
                        &Rewriting  &72     &77\\
    \hline
    \makecell[c]{Objective\\Alignment}  &Package Delivery   &13   &344\\
    \hline
    Test                &Geocoding  &350    &0.4\\
    \hline
    \end{tabular}
    \label{table-time}
\end{table}

\subsubsection{Spatial Encoding in RAG (\textbf{\textit{RQ5}})}
The spatial encoding feature within the RAG module guarantees that the addresses retrieved are spatially close and subsequently influences the information supplied to the LLM.
In this section, we assess the quality of spatial embeddings generated by our retriever in comparison to G2PTL, the most recent state-of-the-art Pre-Trained Model(PTM) for logistics addresses.

\begin{figure}[h]
\centering
\includegraphics[width=\linewidth]{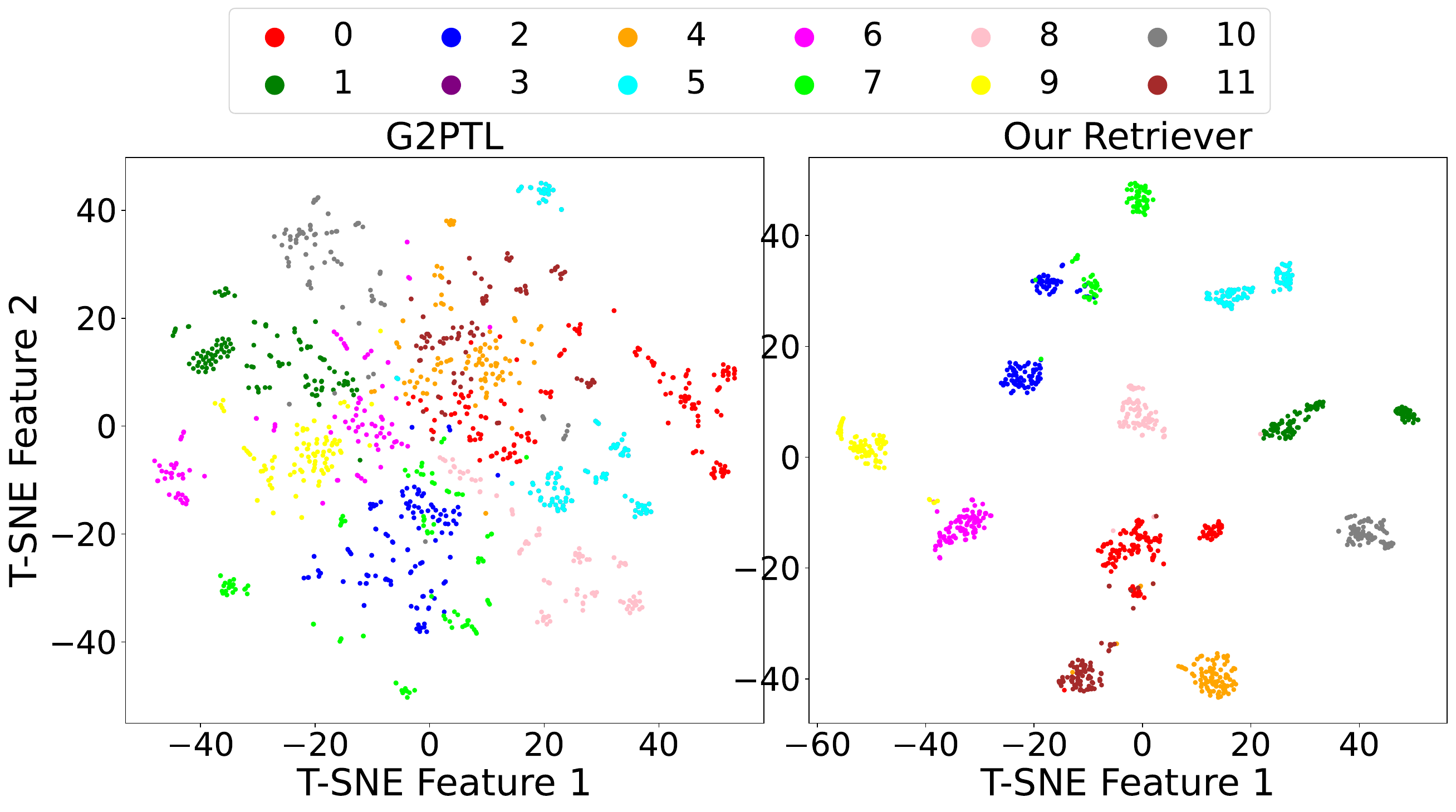}
\caption{T-SNE visualization of address embedding generated by G2PTL and our retriever, from 12 delivery stations in Beijing.}
\label{figure-station}
\end{figure}

Firstly, we select 12 delivery stations across Beijing and utilize the belonged addresses to evaluate model's proficiency in distinguishing between them. We employ t-SNE to reduce dimension and visualize the spatial embedding. 
Station-level classification is at a finer granularity than that of district-level, as a single district may contain dozens of delivery stations.
While G2PTL~\cite{wu2023g2ptl} is reported to have excellent performance in district-level categorization, our observations from Figure~\ref{figure-station} indicate that its capabilities at the station-level do not meet surrounding addresses retrieval in RAG module.
Conversely, our retriever, which has been fine-tuned with a dataset comprising 200 million samples from nationwide geocoding database, demonstrates superior performance in classifying addresses at the station level.

\begin{figure}[h]
\centering
    \begin{subfigure}[b]{0.48\linewidth}
        \centering
        \includegraphics[height=3cm]{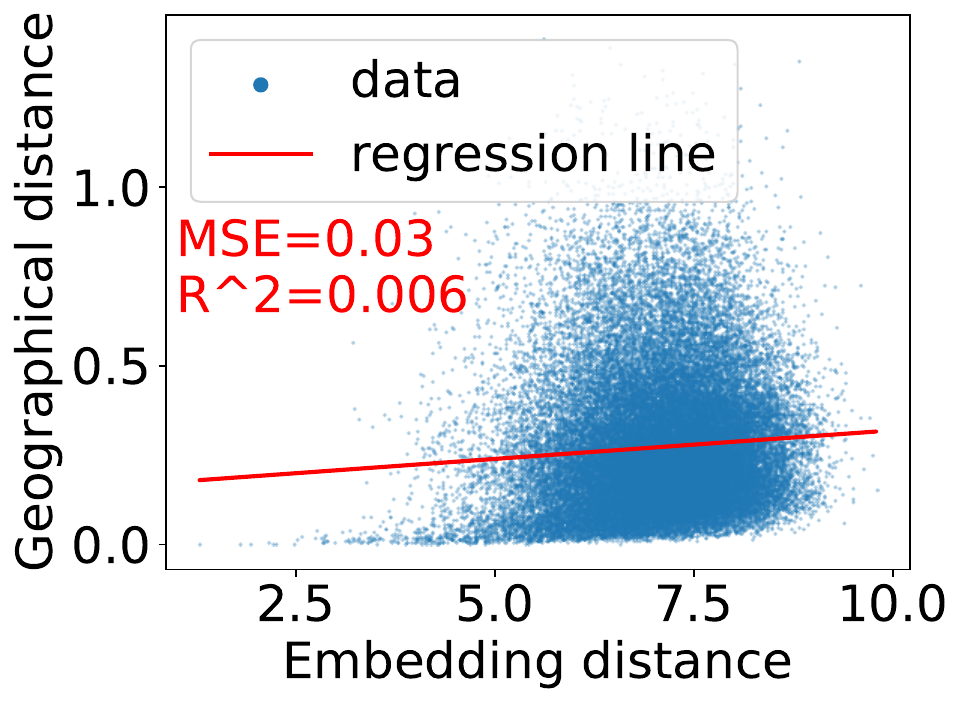}
        \caption{G2PTL}
        \label{subfigure-embedding}
    \end{subfigure}
    \hspace{0.1cm}
    \begin{subfigure}[b]{0.48\linewidth}
        \centering
        \includegraphics[height=3cm]{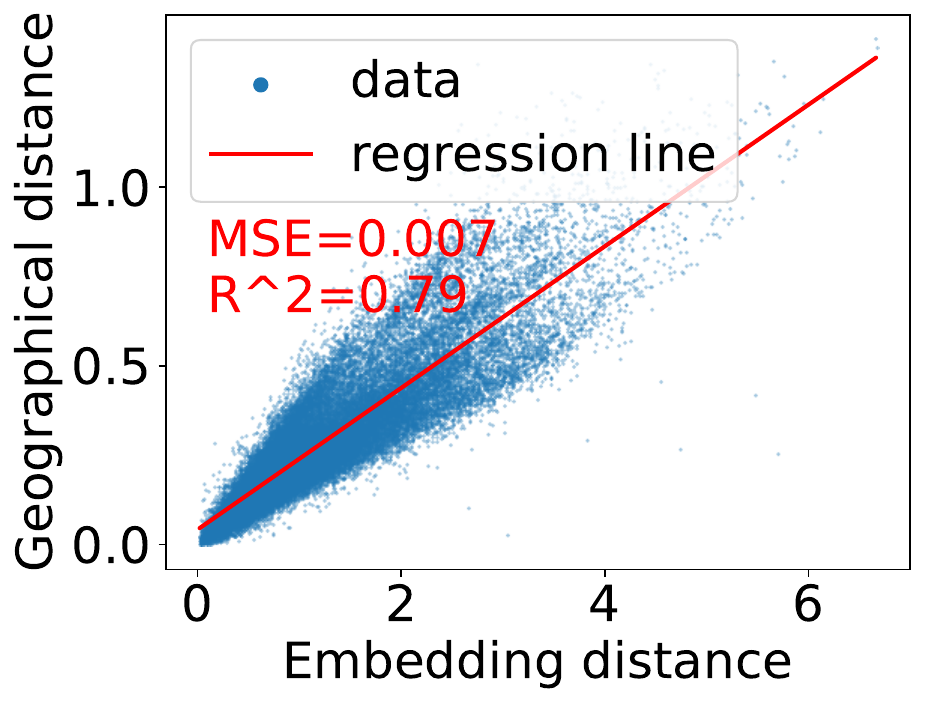}
        \caption{Our retriever}
        \label{subfigure-embedding-finetuned}
    \end{subfigure}
\caption{Embedding visualization of 50000 address pairs in Beijing}
\label{figure-embedding}
\end{figure}

Subsequently, we evaluate the correlation between spatial embedding distance and geographical distance. We collect 50,000 address pairs from Beijing and chart the embedding distance and corresponding geographical distance. A linear regression analysis is then applied to the charted data. We further quantify the correlation by Mean Squared Error(MSE) and coefficient of determination($R^2$). The result is depicted on Figure~\ref{figure-embedding}. The findings indicate that there is a linear correlation between the spatial embedding distances produced by our retriever and the actual geographical distances, demonstrating that our retriever effectively encodes addresses in a way that preserves their real-world spatial relationships.

\begin{figure}[h]
    \centering
    \includegraphics[width=\columnwidth]{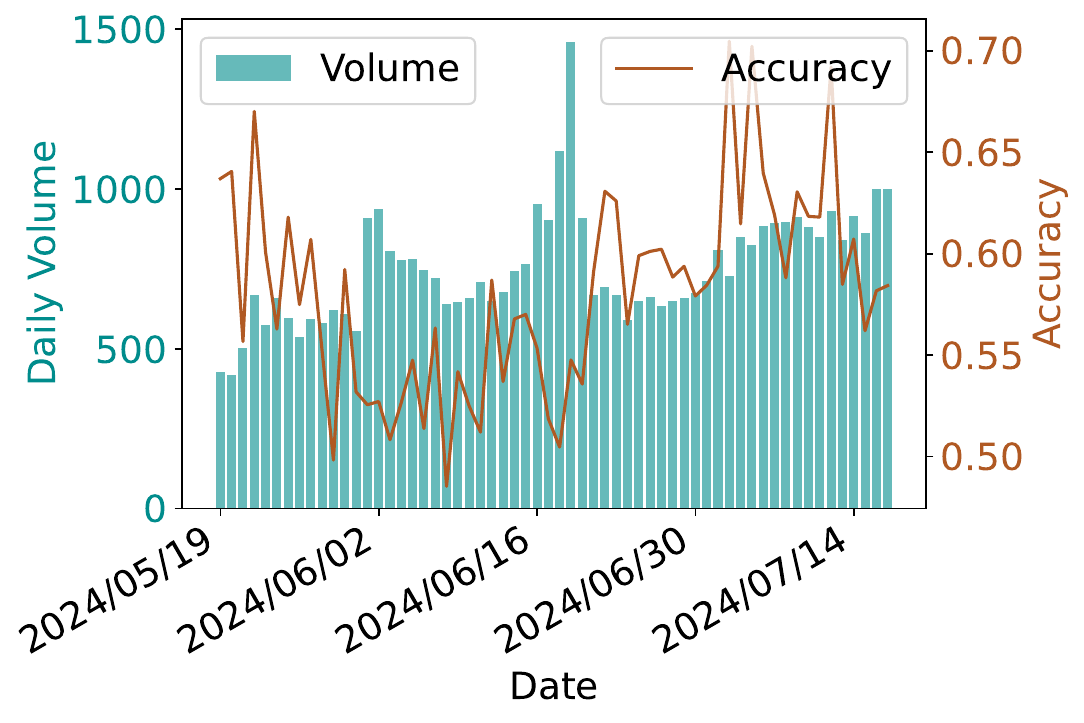}
    \caption{Result of Deployment}
    \label{figure-deployment}
\end{figure}

\subsection{Online Deployment (RQ6)}
AddrLLM is incorporated to JD's current LBS system and deployed in Zhejiang province, China. Specifically, all packages sending from Zhejiang will be processed by the new system. Because the huge computing resources needed by Large Language Models, in the deployed system, AddrLLM is called to rewrite addresses only when abnormal addresses are detected~\cite{hong2022fastaddr}. Considering daily rewriting volume in Zhejiang province, AddrLLM is deployed on a computational cluster with 4 Intel Xeon 8468V CPUs, 4 NVIDIA RTX 4090D GPUs and 256 GB RAM.

We present monitoring data over the course of 60 days encompassing the "618", a sales event comparable to Black Friday. Daily rewriting volume and accuracy (percentage of abnormal addresses that can be corrected by AddrLLM) is depicted in Figure~\ref{figure-deployment}.
On average, there are 754 instances of address rewriting each day. Our specialized address rewriting model, AddrLLM, has effectively corrected over 40\% of these erroneous addresses, ensuring the proper delivery of corresponding parcels. Moreover, the accuracy curve reveals that the performance of AddrLLM remains stable when applied to real-world data streams.

Results from Yulin city, Shaanxi Province and Yangjiang city, Guangdong Province are shown in Appendix~\ref{section-deployment}.
\section{Discussion}

\noindent\textbf{Lessons Learned}:
We summarize key lessons learned from work:
\\\noindent$\bullet$ Large Language Model shows powerful ability of reasoning, but still needs fine-tuning and objective alignment to meet the requirements of specific domain and task.
\\\noindent$\bullet$ Retrieval-Augmented Generation(RAG) boosts LLM's performance in our address rewriting scenario.
\\\noindent$\bullet$ After deployment of LLM, RAG decreases the workload of updating knowledge, especially for a rapidly updating geocoding database.

\noindent\textbf{Limitations}:
\\\noindent$\bullet$ For the offline experiment, because of vast amount of training data required by fine-tuning LLM, it's impractical to obtain a comprehensive analysis of all the incorrect address. For example, in Table~\ref{table-errors}, what are percentages of these errors in real-world address query.
\\\noindent$\bullet$ Because of the black-box essential of LLM, we adopt a relatively conservative way to deploy our model. Instead of directly incorporating our model into core of JD's LBS system, which is the most critical system in a logistics company, we initially utilize it to rectify erroneous addresses after abnormal addresses detected. Though it is a good way to examine model's ability and robustness on real-world scenario, AddrLLM's capabilities is not fully utilized.

\noindent\textbf{Future Work}:
\\\noindent$\bullet$ An autonomous method to analyze different types of errors in daily addresses.
\\\noindent$\bullet$ After obtaining comprehensive analysis of erroneous addresses, train and test LLM's rewriting ability on individual type of error.
\\\noindent$\bullet$ Further examine and enhance LLM's robustness, decrease its inference latency and incorporate it into the core of JD's LBS system.
\section{Related Work}

\subsection{Address Processing}
Address Processing is a critical task in the domain of geospatial analysis, intersecting with Natural Language Processing, information retrieval and Machine Learning. There have been significant researches advancing the field of address-related tasks. ~\cite{roy2023deep} formulated address standardization as a multi-task classification problem and developed a seq-to-seq based low-latency address rewriting framework. ~\cite{hong2022cominer} infers geographic coordinates from textual addresses using data from e-commerce logistics. ~\cite{hong2022fastaddr} develops a framework for quickly detecting abnormal addresses in e-commerce by using a contrastive address augmentation approach and a lightweight attention model. ~\cite{li2023geoglue} introduces GeoGLUE, a benchmark for evaluating geographic natural language understanding. ~\cite{AddrCluster} develops an autonomous method to detect address alias based on logistics data.
Recently, due to the impressive success of Pre-Training Models(PTM), several works~\cite{wu2023g2ptl,huang2022ernie,MGeo} developed PTMs for address related tasks, based on data from logistics and map areas. However, few work has explored large language model's ability in dealing with address-related tasks.

\subsection{Query Rewriting}
Address rewriting can be considered as a specialized form of query rewriting within the context of Geographic Information Systems(GIS). Address rewriting aims to bridge the gap between user-entered addresses and the data indexed within a GIS. Extensive researches on query rewriting have arisen, whose techniques are sometimes applicable to address rewriting. Previous works mainly view query rewriting as a machine translation problem. These works can be divided into: Statistical Machine Translation (SMT)-based methods and Neural Machine Translation (NMT)-based methods. SMT-based methods~\cite{gao2010clickthrough,gao2012learning,gao2012towards,riezler2008translating,riezler2010query} use statistical models to generate translations from original query to rewritten query. Around the mid-2010s, after deep neural network showed its power in feature learning and generalization~\cite{lecun2015deep}, NMT-based methods~\cite{cho2014learning,liu2021geo,roy2023deep} earned more attention. These methods adopt encoder-decoder architecture, where encoder maps original query into hidden representation, and decoder maps representation to rewritten query.
Recently, large language models have demonstrated remarkable capabilities in understanding and generating human-like text~\cite{brown2020language}. They are adopted in query rewriting tasks to enhance search relevance and user experience~\cite{long-tail-taobao,shu2024rewritelm,zhu2024enhancing}. However, these works target at rewriting e-commerce product query and user question, which are different from address in both format and semantics.
There are several works dealing with address rewriting problem~\cite{roy2023deep,liu2021geo}. Compared with our LLM-based methods, these methods have some drawbacks: (1) Substitution and spell correction can only handle a portion of erroneous addresses; (2) Limited knowledge incorporated in the model; (3) When new addresses come, model fine-tuning or re-training is needed.

\subsection{Retrieval Augmented Generation}
Large Language Models exhibits impressive proficiency but grapple with issues like generating inaccurate information and outdated knowledge. Retrieval Augmented Generation(RAG)~\cite{lewis2020retrieval,huang2024survey,gao2023retrieval} presents a promising strategy. This method involves retrieving information from external database prior to generating responses, then utilizing this information as context during the generation process. Subsequent research has explored various modifications to RAG, such as the REALM model~\cite{guu2020retrieval} further refined the relevance of retrieved information through an end-to-end trainable retrieval component. Atlas~\cite{izacard2023atlas} shows retrieval enhancement can compensate for the shortfall in parameter size, by jointly training the retriever and the reader.
Not only does RAG help in improving the factual accuracy of answers, but it also allows LLM to generate texts related to a specific domain, such as Medicine~\cite{kresevic2024optimization}, Biomimicry~\cite{toukmaji2024retrieval} and Electrocardiography~\cite{yu2023zero}, Linguistics~\cite{hu2023rt5}, Music~\cite{jonason2023retrieval}. However, previous studies mainly focus on Question-answering task and retrieve information formatted as natural language paragraphs. Whether RAG is beneficial to address rewriting task and how to retrieve useful information from address database are seldom explored.

\section{Conclusion}
In this paper, we explore address rewriting utilizing retrieval augmented large language model. We design AddrLLM, an RAG-based LLM fine-tuned on several related downstream tasks and objectively aligned with package delivery task.
In the AddrLLM framework, we introduce two pioneering components, namely the RAG for address and Bias-free Objective Alignment, which are seamlessly integrated into the LLM architecture. We subsequently conduct a comprehensive array of empirical investigations on offline real-world dataset and online deployment. Our empirical findings demonstrate that AddrLLM surpasses the performance of existing methods in both rewriting and downstream applications.
On offline experiments, AddrLLM outperforms other baselines and alleviate package re-routing by 43\%. 
During the online deployment phase, AddrLLM operates with stability and effectively corrects over 40\% of the abnormal addresses.

\clearpage
\bibliographystyle{ACM-Reference-Format}
\bibliography{ref}

\clearpage
\appendix

\section{Chinese Address Hierarchy}\label{section-address}
The hierarchy of standard Chinese address is detailed in Table~\ref{table-address}.

\begin{table}[H]
    \centering
    \caption{Chinese Address Hierarchy}
    \begin{tabularx}{0.9\columnwidth}{c|l}
    \hline
    Level&Component\\
    \hline
1      & Province and Autonomous Region \\
2      & City \\
3A     & Administrative district and county \\
3B     & Development zone \\
4A     & Township, street \\
4B     & Village and community \\
4C     & Group and team \\
4D     & Residential area, phase/sector \\
4E     & Road \\
4F     & Directional term \\
5A     & House number \\
5B     & Points of Interest(POI) (e.g., supermarket, hotel) \\
5C     & Building number (e.g., Building No. 35) \\
5D     & Delivery locker (e.g., Fengchao lockers) \\
6A     & Unit and door \\
6B     & Floor \\
6C     & Room number \\
    \hline
    \end{tabularx}
    \label{table-address}
\end{table}

\section{Analysis of erroneous Chinese Addresses}\label{section-error}
We perform manual observation of 10,000 erroneous Chinese addresses, and describe types of error in Table~\ref{table-errors}

\newlength{\leftwidth}
\setlength{\leftwidth}{1.5cm}
\newlength{\rightwidth}
\setlength{\rightwidth}{4.7cm}
\begin{table}[H]
    \centering
    \caption{Error analysis of 10,000 erroneous Chinese addresses.}
    \begin{tabularx}{\columnwidth}{c|c|c}
    \hline
    Error&
    \makecell[c]{Perce\\nt(\%)}
    &Example\\
    \hline
    \begin{tabular}{m{\leftwidth}}
        Missing\\administrative\\region
    \end{tabular}
    &21.3&
    \begin{tabular}{p{\rightwidth}}
        Building 2, JD Headquarters, No. 11 Ke Chuang Street, Beijing
    \end{tabular}\\
    \hline
    \begin{tabular}{m{\leftwidth}}
        Malicious nested address
    \end{tabular}
    &23.2&
    \begin{tabular}{p{\rightwidth}}
        Building 2, JD Headquarters, No. 11 Ke Chuang Street, Daxing District, Beijing, Xuanwu District, Nanjing
    \end{tabular}\\
    \hline
    \begin{tabular}{m{\leftwidth}}
        Alias of\\
        address\\
        entity
    \end{tabular}
    &14.6&
    \begin{tabular}{p{\rightwidth}}
        Building 2, JDH, No. 11 Ke Chuang Street, Daxing District, Beijing
    \end{tabular}\\
    \hline
    \begin{tabular}{m{\leftwidth}}
        Address-irrelevant words
    \end{tabular}
    &27.9&
    \begin{tabular}{p{\rightwidth}}
        Building 2, JD Headquarters, No. 11 Ke Chuang Street, Daxing District, Beijing (Previous China Mobile Station)
    \end{tabular}\\
    \hline
    \begin{tabular}{m{\leftwidth}}
        Misspelling
    \end{tabular}
    &13.1&
    \begin{tabular}{p{\rightwidth}}
        Building 2, DJ Headquarters, No. 11 Ke Chuang Street, Daxing District, Beijing
    \end{tabular}\\
    \hline
    \multicolumn{3}{c}{
    \begin{tabular}{p{0.9\columnwidth}}
        One standard address: Building 2, JD Headquarters, No. 11 Ke Chuang Street, Daxing District, Beijing
    \end{tabular}}\\
    \end{tabularx}
    \label{table-errors}
\end{table}

\section{Prompt}\label{section-prompt}
We demonstrate the prompt for SFT, objective alignment, model testing and online deployment in Table~\ref{table-prompt} and Table~\ref{table-prompt-rewriting}.

\begin{table}[H]
    \centering
    \caption{Prompt for SFT}
    \begin{tabularx}{0.9\columnwidth}{c|c}
    \hline
    Task&Prompt\\
    \hline
    \begin{tabular}{m{1.5cm}}
        \centering Address\\
        \centering Parsing\\
    \end{tabular}
    &
    \begin{tabular}{p{5cm}}
        You are an address parsing bot, please
        parse the following address according
        to standard address hierarchy:\\
        Address:\{address\}\\
        Address Hierarchy:\{address hierarchy\}\\
        System:\{address components\}\\
    \end{tabular}
    \\
    \hline
    \begin{tabular}{m{1.5cm}}
         \centering Address\\
         \centering Entity\\
         \centering Prediction\\
    \end{tabular}
    &
    \begin{tabular}{p{5cm}}
        You are an Address Entity Prediction bot,
        please predict missing address entity in
        the following address:\\
        Address:\{incomplete address\}\\
        Address Hierarchy:\{address hierarchy\}\\
        System:\{address entity\}\\
    \end{tabular}
    \\
    \hline
    \begin{tabular}{m{1.5cm}}
        \centering Address\\
        \centering Rewriting\\
    \end{tabular}
    &
    \begin{tabular}{p{5cm}}
        You are an address rewriting bot, please
        rewrite the following address according
        to standard address hierarchy:\\
        Address:\{address\}\\
        Address Hierarchy:\{address hierarchy\}\\
        Examples:\{address rewriting examples\}\\
        System:\{Rewritten address or original address\}
    \end{tabular}
    \\
    \hline
    \end{tabularx}
    \label{table-prompt}
\end{table}

\begin{table}[H]
    \centering
    \caption{Prompt for Objective Alignment and evaluation}
    \begin{tabularx}{0.9\columnwidth}{p{7cm}}
    \hline
    You are an address rewriting bot, please
    rewrite the following address according
    to standard address hierarchy and related addresses. Related addresses are possibly geographically close to the address to be rewritten. If no rewrite is needed, output the original address \\
    Address to be rewritten:\{address\}\\
    Address Hierarchy:\{address hierarchy\}\\
    Examples:\{address rewriting examples\}\\
    Related Address:\{addresses returned by retriever\}\\
    System: \{rewritten address or original address\}\\
    \hline
    \end{tabularx}
    \label{table-prompt-rewriting}
\end{table}

\section{RL Formulation}\label{section-rl}
In the context of reinforcement learning(RL), the objective alignment of LLM is formulated as a Markov Decision Process(MDP) with the following components $<S,A,r,\pi,\gamma>$.
\\\noindent$\bullet$~\textbf{\textit{State space $S$}}: The state space $S$ is limited by the vocabulary set $V$ and maximum sequence length $L$:
\begin{equation}
    S = \left\{ s \in V^* \big|  |s| \leq L \right\}
\end{equation}
\\\noindent$\bullet$~\textbf{\textit{Action space $A$}}: 
The action space $A$ equals to the vocabulary set V.
\\\noindent$\bullet$~\textbf{\textit{Reward $r$ and Reward function $R$}}: 
Model obtains an immediate reward $r_t$ at time $t$ by a rewarding function $R$. At each time step $t$, policy model $\pi$ generate next token as action $a_t$, based on current state $s_t$.
\\\noindent$\bullet$~\textbf{\textit{Policy model $\pi$ and discount factor $\gamma$}}: 
At time $t$, the model chooses an action following a policy $\pi$:$S\to{A}$. The initial policy model $\pi_0$ is the large language model. At each time step $t$, the model generates the next token, based on current state $s_t$, i.e. already generated tokens. After finishing generation, a reward is calculated by evaluating the output of model.

\section{Deployment Result}\label{section-deployment}
\begin{figure}[H]
    \centering
    \includegraphics[width=\columnwidth]{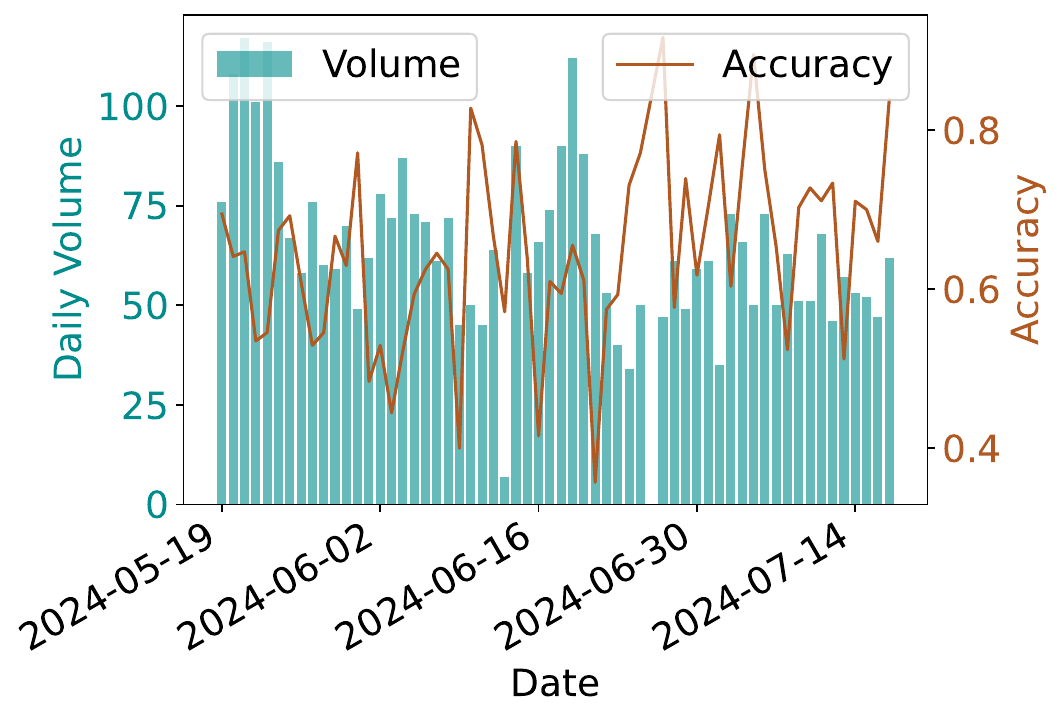}
    \caption{Result of deployment in Yulin}
    \label{figure-yulin}
\end{figure}

\begin{figure}[H]
    \centering
    \includegraphics[width=\columnwidth]{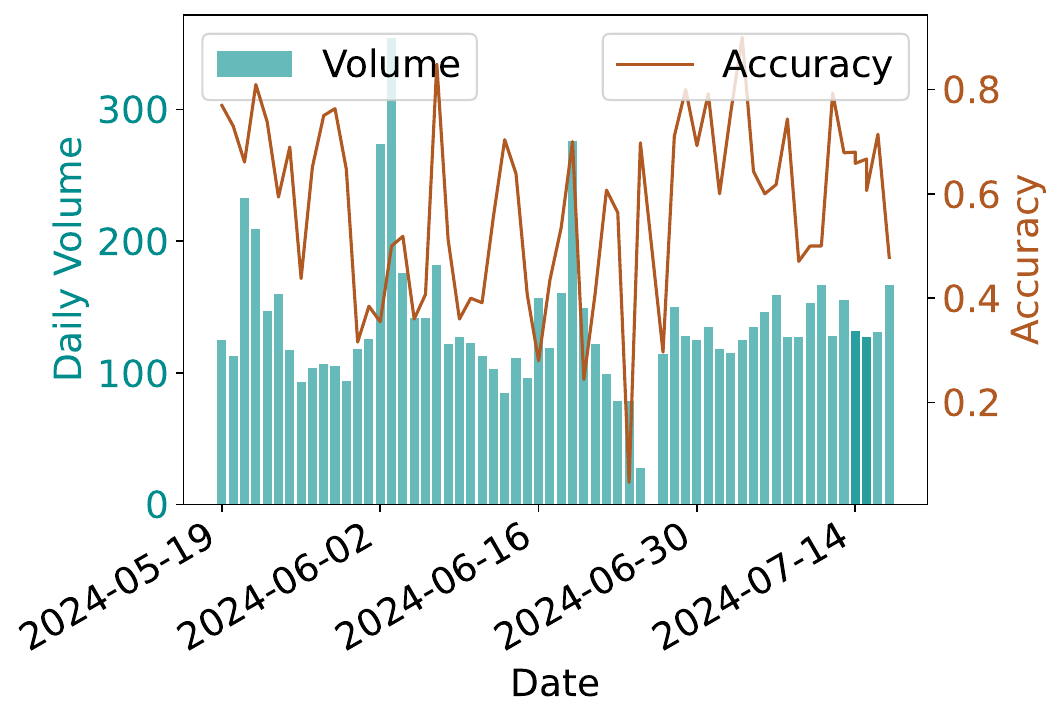}
    \caption{Result of deployment in Yangjiang}
    \label{figure-yangjiang}
\end{figure}

Results from Yulin city, Shaanxi Province and Yangjiang city, Guangdong Province are shown in Figure~\ref{figure-yulin} and Figure~\ref{figure-yangjiang}, respectively.

\end{document}